%% file: main.tex
\documentclass[letterpaper, 10 pt, journal, twoside]{IEEEtran}

\IEEEoverridecommandlockouts    
\usepackage{glossaries}
\input{glossary}

\usepackage{tabularx}
\usepackage{multirow, multicol, makecell}
\usepackage{array}
\usepackage{booktabs}

\usepackage[table,usenames,dvipsnames]{xcolor}      
\usepackage{extarrows}                              
\usepackage{enumitem}
\usepackage{wrapfig}
\usepackage{multirow} 
\usepackage{multicol} 
\usepackage{makecell}
\usepackage{dashrule}
\usepackage{threeparttable}
\usepackage{tablefootnote}
\usepackage{amsmath,amssymb,amsfonts,amsthm,dsfont, bm} 
\usepackage{mathtools}
\usepackage{algorithm,algorithmicx,listings}        
\usepackage[noend]{algpseudocode}			              
\definecolor{myBlue}{RGB}{0,0,0}
\definecolor{ver2blue}{RGB}{0,0,255}

\usepackage{graphicx}
\usepackage{textcomp}
\usepackage[breaklinks=true, colorlinks, bookmarks=true, citecolor=Black, urlcolor=Violet,linkcolor=Black]{hyperref}

\usepackage[normalem]{ulem}
\usepackage[utf8]{inputenc}
\usepackage[english]{babel}
\usepackage{hyperref}
\hypersetup{
    colorlinks=true,
    linkcolor=blue,
    filecolor=magenta,      
    urlcolor=cyan,
}

\usepackage{lipsum}
\usepackage[capitalise]{cleveref}

\usepackage{caption}
\usepackage{graphicx}
\usepackage{multirow}
\usepackage{hhline}

\usepackage{pifont}
\usepackage{graphicx}
\usepackage{tabularx}
\usepackage{arydshln}
\usepackage{hyperref}

\newcolumntype{Y}{>{\centering\arraybackslash}X}

\begin{document}
\renewcommand{\arraystretch}{1.1}
\title{{Neural-Network-Driven Reward Prediction as a Heuristic: Advancing Q-Learning for Mobile Robot Path Planning}}


\author{Yiming Ji, Kaijie Yun, Yang Liu$^{*}$, Zongwu Xie, and Hong Liu
\thanks{$^{*}$corresponding author}%
\thanks{All authors are with State Key Laboratory of Robotics and Systems, Harbin Institute of Technology. (Corresponding author: Yang Liu). {\tt\footnotesize jiyiming@alu.hit.edu.cn,liuyanghit@hit.edu.cn}}%
}

\maketitle

\begin{abstract}

Q-learning is a widely used reinforcement learning technique for solving path planning problems. It primarily involves the interaction between an agent and its environment, enabling the agent to learn an optimal strategy that maximizes cumulative rewards.
Although many studies have reported the effectiveness of Q-learning, it still faces slow convergence issues in practical applications.
To address this issue, we propose the NDR-QL method, which utilizes neural network outputs as heuristic information to accelerate the convergence process of Q-learning.
Specifically, we improved the dual-output neural network model by introducing a start-end channel separation mechanism and enhancing the feature fusion process. After training, the proposed NDR model can output a narrowly focused optimal probability distribution, referred to as the guideline, and a broadly distributed suboptimal distribution, referred to as the region. Subsequently, based on the guideline prediction, we calculate the continuous reward function for the Q-learning method, and based on the region prediction, we initialize the Q-table with a bias.           
We conducted training, validation, and path planning simulation experiments on public datasets. The results indicate that the NDR model outperforms previous methods by up to 5\% in prediction accuracy. Furthermore, the proposed NDR-QL method improves the convergence speed of the baseline Q-learning method by 90\% and also surpasses the previously improved Q-learning methods in path quality metrics.

\begin{IEEEkeywords}
    Q-learning algorithm; Reinforcement learning; Neural network; Path planning.
\end{IEEEkeywords}

\end{abstract}
     \input{secs/introduction.tex}
     \input{secs/related_work.tex}

     \input{secs/method.tex}

     \input{secs/experiments.tex}

     \input{secs/conclusion.tex}
\bibliography{main}

\end{document}

%% file: glossary.tex
\newacronym{mav}{MAV}{Micro Aerial Vehicles}
\newacronym{uav}{UAV}{Unmanned Aerial Vehicle}
\newacronym{ovc}{OVC}{Open Vision Computer}
\newacronym{lidar}{LiDAR}{Light Detection and Ranging}
\newacronym{vio}{VIO}{Visual-Inertial Odometry}
\newacronym{gpgpu}{GPGPU}{General-Purpose Graphics Processing Unit}
\newacronym{ugv}{UGV}{Unmanned Ground Vehicle}
\newacronym{uwb}{UWB}{Ultra Wideband}
\newacronym{svm}{SVM}{Support Vector Machine}
\newacronym{fcn}{FCN}{Fully Convolutional Network}
\newacronym{cnn}{CNN}{Convolutional Neural Network}
\newacronym{loam}{LOAM}{LiDAR Odometry and Mapping}
\newacronym{sloam}{SLOAM}{Semantic LiDAR Odometry and Mapping}
\newacronym{slam}{SLAM}{Simultaneous Localization and Mapping}
\newacronym{iot4ag}{IoT4Ag}{NSF Engineering Research Center for the Internet of Things for Precision Agriculture}
\newacronym{grasp-lab}{GRASP Lab}{the General Robotics, Automation, Sensing and Perception Laboratory}
\newacronym{jps}{JPS}{Jump Point Search}
\newacronym{ukf}{UKF}{Unscented Kalman Filter}
\newacronym{sam}{SAM}{Smoothing and Mapping}
\newacronym{slc}{SLC}{Semantic Loop Closure}
\newacronym{icp}{ICP}{Iterative Closest Point}
\newacronym{imu}{IMU}{Inertial Measurement Unit}
\newacronym{tsdf}{TSDF}{Truncated Signed Distance Field}
\newacronym{esdf}{ESDF}{Euclidean Signed Distance Field}
\newacronym{sdf}{SDF}{Signed Distance Field}
\newacronym{rrt}{RRT}{Rapidly Exploring Random Tree}
\newacronym{fpv}{FPV}{First-person View}
\newacronym{dnn}{DNN}{Deep Neural Network}
\newacronym{igpred}{IGPred}{Information Gain Prediction}
\newacronym{csqmi}{CSQMI}{Cauchy-Schwarz Quadratic Mutual Information}
\newacronym{nbv}{NBV}{Next Best View}
\newacronym{vae}{VAE}{Variational Autoencoder}
\newacronym{tsp}{TSP}{Travelling Salesman Problem}
\newacronym{bgsm}{BGSM}{Behavior Guidance State Machine}
\newacronym{pca}{PCA}{Principal Component Analysis}
\newacronym{aspp}{ASPP}{Atrous Spatial Pyramid Pooling}
\newacronym{swap}{SWaP}{Size Weight and Power}
\newacronym{soi}{SoI}{Semantic Object of Interest}
\newacronym{aoi}{AoI}{Area of Interest}
\newacronym{cop}{COP}{Correlated Orienteering Problem}
\newacronym{ig}{IG}{Information Gain}

%% file: secs/introduction.tex
\section{Introduction}
\label{sec:intro}
Path planning is  fundamental for autonomous robots to move efficiently between locations. A good path must be free of collisions and enable the robot to reach its destination quickly using the shortest distance \cite{bai2024path}.
Traditional path planning methods include algorithms such as Rapidly-exploring Random Trees (RRT), Probabilistic Roadmaps (PRM), Artificial Potential Fields (APF), A*, and Visibility Graphs (VG). These approaches can be categorized into graph-based, data-based, and sampling-based methods.

Graph-based methods like A* efficiently find the shortest path but suffer from exponential increases in computational complexity as the search space expands, making them unsuitable for environments with complex and continuous obstacles \cite{liao2024research}. Data-based methods, such as the APF algorithm \cite{zhang2024agv}, tend to encounter numerous zero-potential energy points, which can cause the algorithm to become trapped in local minima and fail to generate a complete path.
Sampling-based methods like RRT leverage random sampling for path planning \cite{zhang2024intelligent}, offering robust search capabilities and rapid exploration speeds. However, they are limited by lower search accuracy and poorer path smoothness compared to other methods. 

In recent years, learning-based technology has grown into a critical component of autonomous mobile robotics, especially in the area of path planning. While classical path planning methods have long struggled with certain inherent limitations, recent developments in AI-driven solutions have begun to overcome these challenges. Among these approaches, Q-learning (QL) is particularly promising. Its model-free nature and ability to incrementally learn from ongoing interactions with the environment make it well-suited for addressing the complex optimization problems that arise when determining efficient, reliable paths for autonomous robots \cite{haghzad2017path}.

Q-learning fundamentally operates on the idea that each action taken in a given state is either rewarded or penalized, prompting the algorithm to seek a policy that maximizes long-term returns \cite{watkins1989learning}.
To accomplish this, QL relies on a Q-table—a lookup structure that associates every state-action pair with its expected payoff, known as a Q-value. By repeatedly exploring and updating these values within the environment, QL can, in theory, converge toward an optimal decision-making policy \cite{prince2023understanding}.
\begin{figure}[t]
    \begin{center}
        \includegraphics[width=\linewidth]{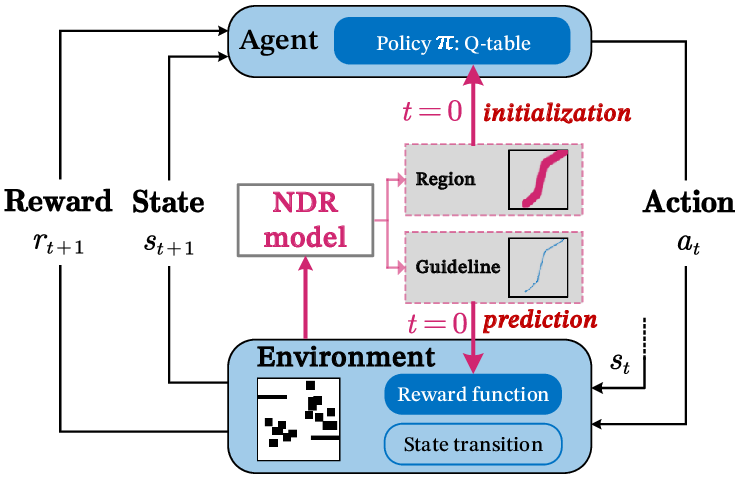}
    \end{center}
    \caption{
        The proposed NDR-QL method. At $t = 0$, the pre-trained NDR model is invoked to initialize each Q-table and predict the continuous reward function.
        }
    \label{fig:fig7}
\end{figure}
However, a significant limitation arises from the curse of dimensionality, which slowing down the learning process. As the state space or the range of possible actions grows larger, the size of the Q-table increases exponentially, creating substantial memory demands \cite{vanhulsel2009simulation}. This not only restricts QL’s practicality for real-world physical systems but also impedes efficient learning. Moreover, traditional QL typically initializes Q-values uniformly or at random, offering no initial insight into the environment and further slowing the rate of convergence \cite{xu2019path}.
To this end, this study proposes an improved QL method aimed at accelerating the algorithm’s convergence. Building upon conclusions drawn from previous research \cite{huang2024efficiency}, we redesigned the Neural-Network-Driven Prediction model, denoted as the NDR model, and employed its guideline and region predictions as prior information to guide the QL algorithm, thereby achieving acceleration. The overall framework is illustrated in Fig. \ref{fig:fig7}.

The contributions of this paper are summarized as follows:
\begin{itemize}
    \item We have improved the CNN-based path region prediction model by separating the start and end points from the map channels and introducing an \textit{attention fusion module} to integrate high-level and low-level features. Training and validation on public datasets demonstrate that our proposed NDR model outperforms the previous state-of-the-art models.
    
    \item We propose two methods that leverage the information output by the pre-trained NDR model to inform QL, thereby accelerating the convergence of the algorithm. The first method is the \textit{Guideline-based Reward Prediction}, which uses the guidelines predicted by the NDR model to compute the continuous reward function of the map. The second method is the \textit{Region-based Q-table Initialization}, which initializes the Q-table using the regions output.
    
    \item Experimental results demonstrate that our proposed NDR-QL method outperforms previously enhanced QL approaches in both convergence speed and path optimality. Furthermore, our method enables the agent to navigate more directly toward the endpoint, reduces the randomness of initial exploration, and accelerates the agent’s perception of the environment.
\end{itemize}

%% file: secs/related_work.tex
\section{Related Work}
\subsection{Neural-network-driven prediction}



With the advancement of AI technology, a novel approach integrating neural networks with path planners has emerged. These neural network models are trained on various planning scenarios to identify promising regions, which then serve as sampling domains for non-uniform samplers \cite{li2023lightweight}. By ensuring that these regions encompass the optimal path, samples are preferentially generated near this path, thereby accelerating the search process. For example, algorithms such as NRRT* \cite{wang2020neural} and RGP-RRT* \cite{huang2024efficiency} employ CNN-based models to predict promising areas, while MPT \cite{johnson2021motion} utilizes transformer-based models to identify specific regions on maps. Neural-network-driven prediction models have significantly enhanced the performance of traditional sampling-based methods. Moreover, we propose that the global prior heuristic information provided by these models can also address the slow convergence issues associated with Q-learning methods.

\subsection{Q-learning}
In path planning tasks, traditional QL algorithms encounter slow convergence issues.
Therefore, a lot of scholars\cite{low2019solving,maoudj2020optimal,low2023modified,zhou2024optimized} proposed methods to improve the efficiency and performance of Q-Learning. Low et al.\cite{low2019solving} introduced the Flower Pollination Algorithm (FPA) to QL and convergence is obviously accelerated when Q-values are initialized properly by FPA. Maoudj et al.\cite{maoudj2020optimal} proposed an Efficient Q-Learning (EQL) algorithm which introduce a new reward function to initialize the Q-table and provide prior environmental knowledge, alongside a novel selection strategy to accelerate learning by reducing the search space and ensuring rapid convergence to an optimized solution. Zhou et al.\cite{zhou2024optimized} adopted the similar idea presented Optimized Q-Learning (O-QL).

%% file: secs/method.tex
\section{Methodology}
\label{sec:method}
\subsection{Network Structure}
\label{subsec:network}
The previous methods (RGP \cite{huang2024efficiency}, NRRT* \cite{wang2020neural}, NEED \cite{wang2021robot}) all use 3-channel RGB as input, with the map represented by black-and-white pixels and the start and end positions marked by blue and red pixels. 
In contrast, the proposed NDR model separates the start point, end point, and map into three distinct input channels. The NDR model uses the STDC \cite{fan2021rethinking} backbone instead of the traditional ResNet \cite{he2016deep}.

\begin{figure*}[htbp]
    \begin{center}
        \includegraphics[width= 0.9\linewidth]{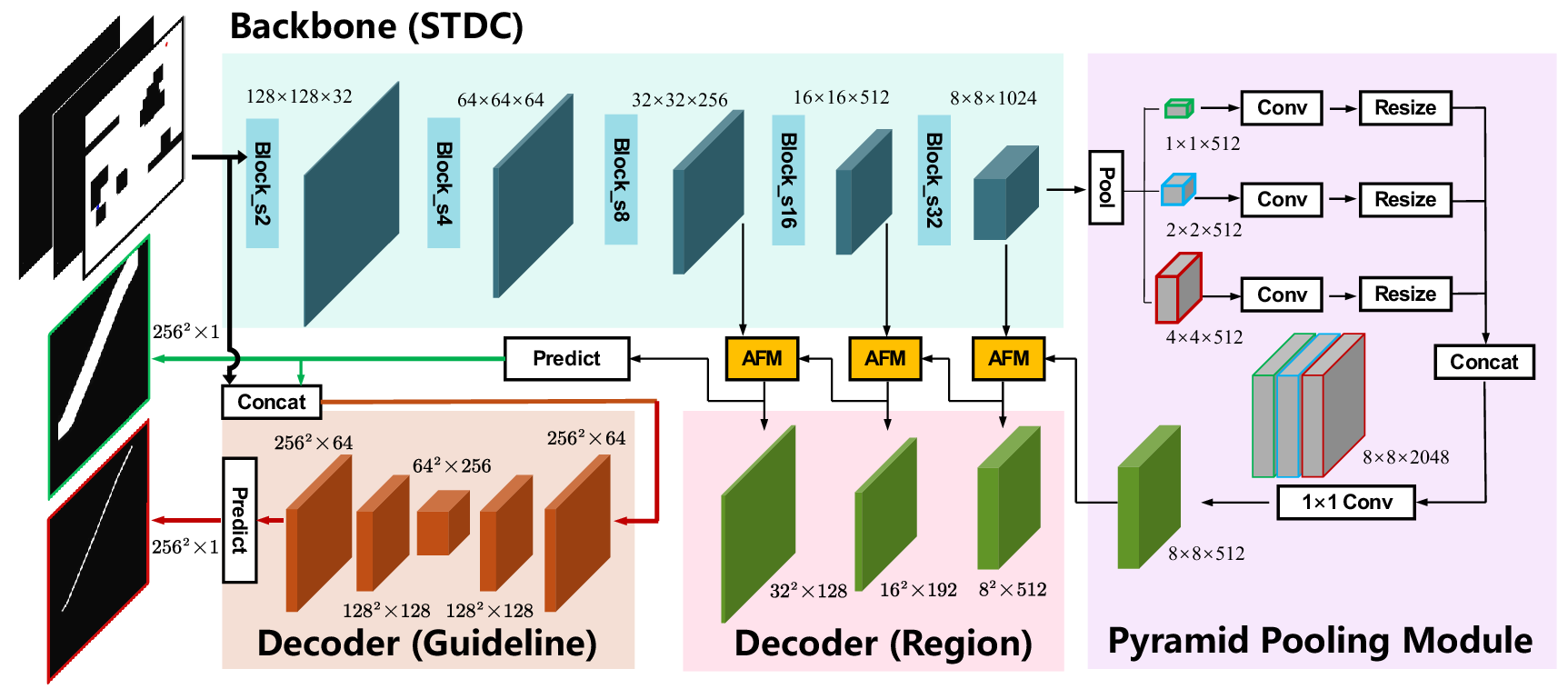}
    \end{center}
    \caption{
        Structure of the NDR model.
        }
    \label{fig:fig1}
\end{figure*}
Within its multi-stage encoding architecture, the spatial resolution of feature maps is reduced by a factor of two after each STDC stage. As such, a high-level semantic description of the input map can be extracted as the final stage5 output of the STDC backbone. The low-level features output by stage3 and stage4 will be fused with the high-level features in the decoder to enhance the model’s performance for path planning.
As NRRT* has highlighted, fusing features from multiple semantic levels can significantly enhance performance in tasks demanding precise localization, such as path planning. Therefore, we employ AFM \cite{peng2022pp} modules that integrate both low-level spatial details and high-level semantic information.

To enhance global scene understanding, PSPNet \cite{zhao2017pyramid} uses a pyramid pooling module (PPM) that combines pooled features at multiple scales before the convolution layers. This creates both local detail features and global context representations. We simplified the original PPM by using three pyramid scales instead of four and removing the concatenation between the original feature map and upsampled features. This modification reduces computational costs while maintaining effectiveness. 

The RGP model is designed with two separate decoders for predicting regions and guidelines. Following this setup, we concatenate the region prediction with the initial map information and feed this combined input into a four-layer UNet decoder \cite{ronneberger2015u}, ultimately producing the guideline prediction.

In our ablation studies, we further investigate the impact of separating the start/end channels and incorporating positional encoding on the model’s performance.

\subsection{Guideline-based Reward Prediction}
\label{subsec:reward}
In RL-based path planning, an agent aims to maximize its cumulative reward by interacting with the environment as it moves from initial to target state. The common design of the reward function is as shown in Eq. \ref{eq:eq1}:

\begin{equation}  
    \label{eq:eq1}
    {R=\left\{ \begin{array}{c}
        -10,scenario1\\
        40,scenario2\\
        -5,scenario3\\
        r_c,scenario4\\
        \end{array} \right.}
\end{equation}
Here, $scenario1$ is collision with obstacles or boundary violation, $scenario2$ is reaching the target, $scenario3$ is revisiting a node, and $scenario4$ covers all other cases with reward $r_c$.

A key challenge in RL-based path planning is the sparse reward problem. To address this, continuous reward functions were introduced. These functions provide immediate feedback at each step instead of only at completion. This helps agents learn action impacts and optimize their strategies more effectively \cite{zhang2021novel}. The smooth nature of continuous rewards, compared to discrete ones, better represents environmental changes and helps predict rewards for unexplored states.

The conventional \textit{Distance-based Continuous Reward Function (D-CRF)}, is defined as in Eq. \ref{eq:eq2}:
\begin{equation}  
    \label{eq:eq2}
    {r_c = r_0 + r_{\max} \times \exp\biggl(-(\frac{x_i - x_d}{G_x}) - (\frac{y_i - y_d}{G_y})\biggr)}
\end{equation}
Where $r_0$ is the base reward for moving into empty space and $r_{max}$ is the maximum reward at the target. $G_x$ and $G_y$ respectively control the decrease rate of the reward along the x-axis and y-axis.

This paper proposes a \textit{Neural Network Prediction-based Continuous Reward Function (NDR-CRF)}, which utilizes the guideline predictions (denoted as ${NDR}_g$) generated by the NDR model introduced in Section \ref{subsec:network} to compute the reward values associated with each state in the map, as in Eq. \ref{eq:eq3}.
\begin{equation}  
    \label{eq:eq3}
    {r_c\left( x_i,y_i \right) =NDR_g\left( x_i,y_i \right) \times r_{\max}}
\end{equation}

Using $NDR_g$ to predict rewards is reasonable because, ideally, the NDR model can identify an optimal path region. In RL-based path planning, designing the reward function so that the agent receives higher rewards upon reaching this optimal region and lower rewards when deviating from it allows the agent to leverage prior knowledge rather than starting with no information. 
Subsequent experiments show that \textit{NDR-CRF} enables faster convergence in Q-learning compared to \textit{D-CRF}.
\begin{figure}[htbp]
    \begin{center}
        \includegraphics[width= \linewidth]{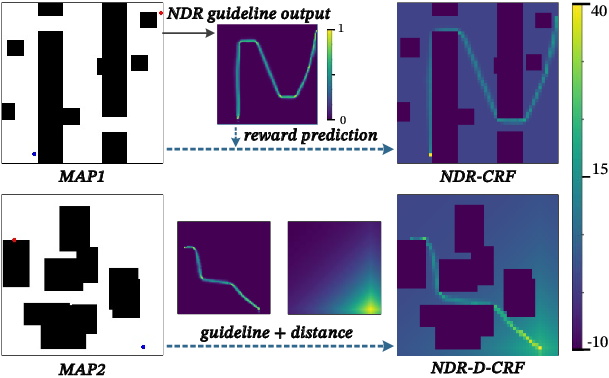}
    \end{center}
    \caption{
        A schematic visualization of the reward function.
        }
    \label{fig:fig2}
\end{figure}

As illustrated in the Fig. \ref{fig:fig2}, before initiating the RL algorithm’s iterative process, we first use the NDR model’s guideline predictions (as defined by Eq. \ref{eq:eq3}) to pre-compute the reward function distribution (NDR-CRF) across the entire map. During subsequent RL iterations, querying this distribution for a specific state provides the corresponding reward value at that step. 
Furthermore, this approach can be combined with the conventional D-CRF method to achieve even stronger performance, as in Eq. \ref{eq:eq4}.
\begin{equation}
    \label{eq:eq4}
    \begin{aligned}
    &r_c(x_i,y_i) = \omega \bigl[ NDR_g(x_i,y_i) \times r_{\max} \bigr] \\
    &\quad + (1-\omega) \biggl[ r_{\max} \times \exp\biggl(-\frac{x_i - x_d}{G_x} - \frac{y_i - y_d}{G_y}\biggr) \biggr]
\end{aligned}
\end{equation}

Where $\omega$ is the weighting coefficient for the two methods, and for simplicity, $r_0$ is omitted.

\subsection{Region-based Q-table Initialization}
\label{subsec:q-table}
\begin{figure}[t]
    \begin{center}
        \includegraphics[width= \linewidth]{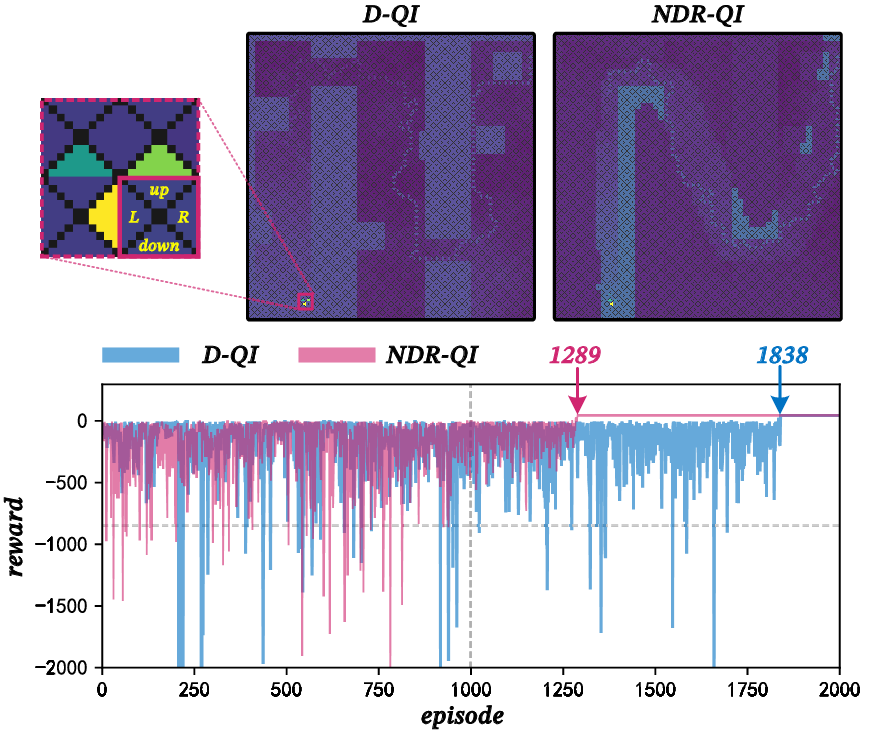}
    \end{center}
    \caption{
        A schematic visualization of the Q-table and a line chart illustrating reward variations during QL algorithm iterations. This experiment only examines the influence of different Q-table initialization methods on convergence speed, using a fixed learning rate and a fixed $\varepsilon$  parameter, without employing any reward prediction techniques.
        }
    \label{fig:fig3}
\end{figure}
In the initial learning phase, the agent knows little about the environment and randomly selects actions. To speed up the convergence of the Q-Learning algorithm, initializing the Q-table with prior knowledge is effective. The previous method \cite{zhou2024optimized} used Euclidean distance between the mobile robot's current and target position, as in Eq. \ref{eq:eq5} , to initialize the Q-table.
\begin{equation}  
    \label{eq:eq5}
    {Q\left( \left[ x_i,y_i \right] \right) =\exp \left( -\sqrt{\left( x_d-x_i \right) ^2+\left( y_d-y_i \right) ^2} \right)}
\end{equation}

The \textit{distance-based Q-table initialization method (referred to as D-QI)} guides the agent toward the target by consistently favoring the action with the highest Q-value in each state. However, this method leverages only limited prior knowledge and its initial Q-values are not tailored to the specific environment.
In this work, we introduce a Q-table initialization strategy utilizing region predictions generated by the NDR model, hereafter denoted as NDR-QI, as defined in Eq. \ref{eq:eq6}:
\begin{equation}
    \label{eq:eq6}
    \begin{aligned}
    &Q\left( \left[ x_i,y_i \right] \right) =\omega \cdot MASK\left[ N_r\left( x_i,y_i \right) \right] \\
    &\quad + \left( 1-\omega \right) \cdot \exp \left( -\sqrt{\left( x_d-x_i \right) ^2+\left( y_d-y_i \right) ^2} \right) 
\end{aligned}
\end{equation}
Where $N_r\left( x_i,y_i \right)$ is the region prediction generated by the NDR model, and $MASK\left[ N_r\left( x_i,y_i \right) \right]$ is a binary mask that assigns a value of 0 to regions with high prediction values and -10 to the rest, as in Eq. \ref{eq:eq7}. The weighting coefficient $\omega$ controls the balance between the two initialization methods.
\begin{equation}  
    \label{eq:eq7}
    {MASK\left[ N_r\left( x_i,y_i \right) \right] =\begin{cases}
        0, \mathrm{if}     N_r\left( x_i,y_i \right) >thd\\
        -10, \mathrm{else}\\
        \end{cases}}
\end{equation}

The rationale for employing the region prediction generated by the NDR model as a condition for Q-table initialization is that, during training, we observed stronger connectivity in region predictions compared to guideline predictions. In this context, connectivity refers to the ability of the model’s sigmoid-based output region to continuously link the start and end points. 
Thus, even if the NDR model’s guideline prediction outputs attain a high F1 score, insufficient connectivity will lead to suboptimal results when using these outputs as conditions for Q-table initialization.
As shown in the unseen sample in Fig. \ref{fig:fig4}-(b), the guideline prediction failed to connect, whereas the region covered a larger area, successfully linking the start and end points.

To determine the threshold $thd$ in Eq. \ref{eq:eq7}, we adopt an adaptive procedure. Initially, we set $thd=0.99$. If the binarized segmentation does not connect the designated start and end points, we decrement $thd$ by 0.01 and reassess connectivity. By iterating this process, we ultimately obtain an optimal threshold that ensures connectivity between the specified points.

We visualize the Q-table by partitioning the map grid along its diagonal into four distinct quadrants, each corresponding to one of the agent’s four actions: \textit{moving forward, moving backward, turning left, and turning right}, as illustrated in Fig. \ref{fig:fig3}.
Compared to the distance-based Q-table initialization method (D-QI), the proposed NDR-QI approach—which employs the region prediction outputs of the NDR model for initialization—effectively restricts the agent’s interactions to a specified area of the environment. Naturally, due to the influence of the $\varepsilon$-greedy exploration strategy, the agent’s exploration occasionally extends beyond these designated regions. Furthermore, as evidenced by the reward convergence trends over training iterations, D-QI converges only after 1838 episodes, whereas NDR-QI reduces this number to 1289 episodes, thereby demonstrating its enhanced efficiency and overall effectiveness.

%% file: secs/experiments.tex
\section{Experiment}
\subsection{Experimental setup}
We utilized modified publicly datasets\footnote{\href{https://github.com/JimOriginal/path-planning-dataset}{https://github.com/JimOriginal/path-planning-dataset}} to validate the effectiveness of the proposed prediction model, NDR, as well as the advanced Q-learning approach informed by its predictive outputs. All model training and inference were conducted on an NVIDIA RTX 4090 GPU. To construct the Q-learning experimental environment, the original $201\times201$ map was downscaled to a $50\times50$ version.

\subsection{Prediction Results}

We evaluate the NDR model in comparison with other state-of-the-art prediction models, including RGP \cite{huang2024efficiency}, NEED \cite{wang2021robot}, and MPT \cite{johnson2021motion}, in terms of region prediction accuracy on both seen and unseen datasets.
In accordance with the RGP framework, we employ the F1 score to assess the prediction accuracy of the model.
The validation results are presented in Tab. \ref{tab:tab1}.
Compared to the previous state-of-the-art method RGP, the NDR model improved the F1 score for identifying guidelines by 5\% on seen datasets and by 4\% on unseen datasets.
We attribute this improvement to the input maps that separates the start and end channels and the introduction of the Attention Fusion Module (AFM) in the NDR model.
Some prediction examples in the validation set are shown in Fig. \ref{fig:fig4}.
Notably, the seen datasets closely resemble the training data, whereas the unseen datasets differ significantly from the training data.

\begin{table}[htbp]
    \centering
    \setlength{\tabcolsep}{4pt}  
    \begin{tabularx}{\columnwidth}{l : *{4}{Y}} 
    \hline\hline
    \multirow{2}{*}{Models} & \multicolumn{2}{c}{Seen Scenario}   & \multicolumn{2}{c}{Unseen Scenario}  \\ 
    \cline{2-5}
                            & Region           & Guideline        & Region           & Guideline         \\ 
    \hline
    RGP \cite{huang2024efficiency}                     & 92.81\%          & 58.16\%          & 59.74\%          & 35.82\%           \\
    NEED \cite{wang2021robot}                    & 92.34\%          & -                & 59.16\%          & -                 \\
    MPT \cite{johnson2021motion}                     & 87.70\%          & -                & 49.75\%          & -                 \\ 
    \hdashline[1pt/1pt]
    NDR                     & \textbf{93.02\%} & \textbf{61.22\%} & \textbf{71.96\%} & \textbf{37.31\%}  \\
    \hline\hline
    \end{tabularx}
    \caption{Comparison experiments with different Neural-Network-Driven Prediction models, evaluated on the validation set of a standard dataset, using the same F1 score metrics as in previous work.}
    \label{tab:tab1}
\end{table}
\begin{figure}[t]
    \begin{center}
        \includegraphics[width= \linewidth]{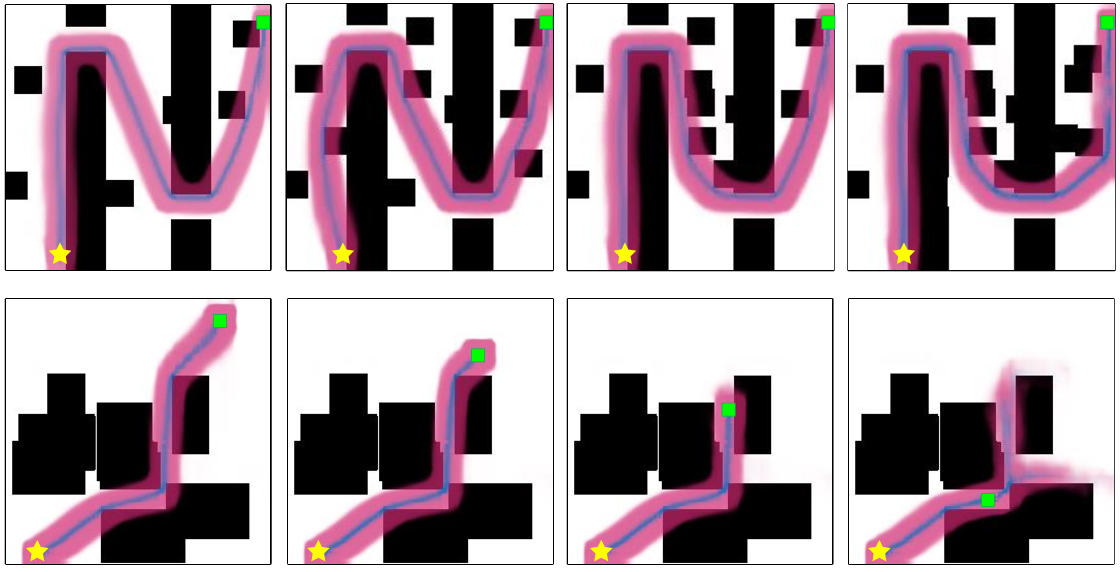}
    \end{center}
    \caption{
        Results from experiments in dynamic scenarios. A green square marks the starting point, and a yellow star denotes the endpoint. In the top row, from left to right, the start and endpoint remain fixed while the environment changes dynamically. In the bottom row, from left to right, we simulate the agent continuously moving along the predicted path.
        }
    \label{fig:fig5}
\end{figure}
To validate the capability of the NDR model in online RL-based planners, we designed two dynamic scenarios. The first scenario involves fixed start and end points while altering the distribution of obstacles on the map, as shown in the first row of Fig. \ref{fig:fig5}.
The second scenario simulates the agent's progression, with a fixed endpoint and continuously changing start points, as depicted in the second row of Fig. \ref{fig:fig5}.
It can be observed that the NDR model accurately predicts the optimal path's regions and guidelines in complex dynamic scenarios, thereby laying the foundation for its application in enhancing the Q-learning algorithm presented in this paper.
\begin{figure}[htbp]
    \begin{center}
        \includegraphics[width= \linewidth]{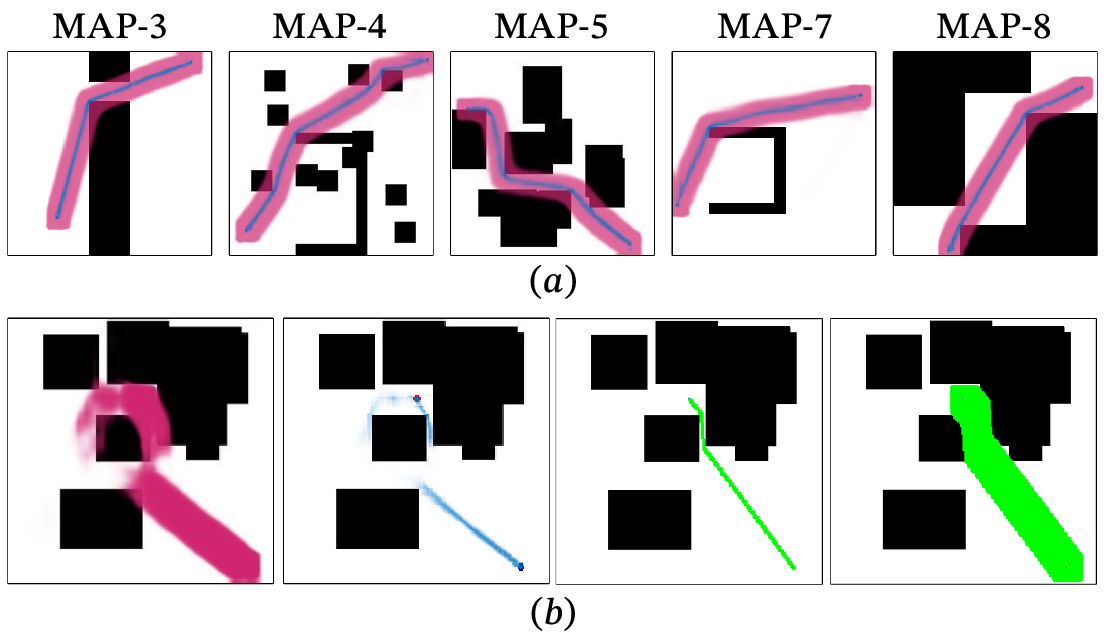}
    \end{center}
    \caption{
        Examples of NDR model predictions. (a) Five map examples; (b) Examples with poor performance. From left to right: region prediction, guideline prediction, guideline and region ground truths.
        }
    \label{fig:fig4}
    \end{figure}
\begin{figure*}[t]
    \begin{center}
        \includegraphics[width=\linewidth]{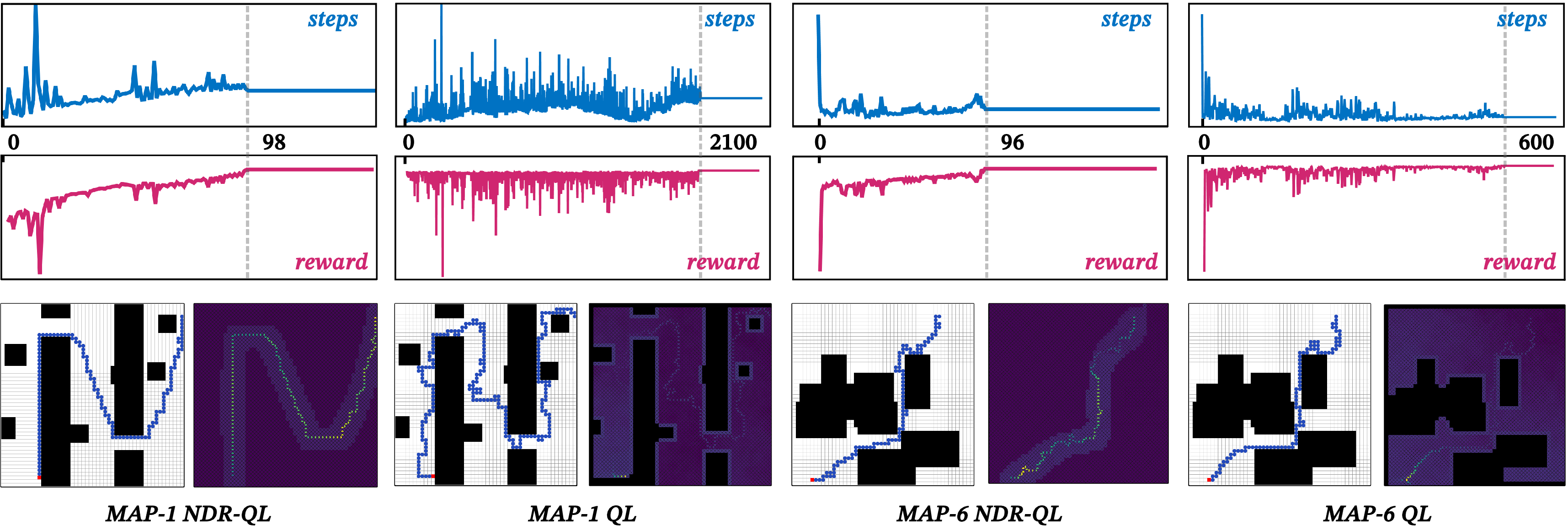}
    \end{center}
    \caption{
        Computational results of the proposed NDR-QL algorithm on several map samples, and its comparison with traditional QL methods.
        }
    \label{fig:fig6}
\end{figure*}

\subsection{Path Planning Results and Comparative Analysis}
In this subsection, we compare the proposed NDR-QL method with O-QL \cite{zhou2024optimized}, IQL \cite{low2022modified}, and the basic QL method across eight map samples, as presented in Tab. \ref{tab:tab3}.
Notably, we employ the distance-based Q-table initialization method and the continuous reward function proposed by IQL, without adopting its improved learning rate adjustment strategy and enhanced action-selection policy. This focus allows us to evaluate the performance improvements brought by different heuristic priors to QL.
The experiments in Tab. \ref{tab:tab3} uniformly employ a fixed learning rate of 0.1 and a fixed $\varepsilon = 0.2$ for the greedy method.
In the table, ``convergence steps" represent the cumulative steps of the QL algorithm from episode 0 until convergence. Smaller convergence steps indicate that the algorithm converges more rapidly.
The table also records the shortest and longest distances of the valid paths obtained during the algorithm's solution process, which serve as metrics for evaluating path quality.

It can be observed that the NDR-QL method improves the convergence speed of the QL method by an average of 90\%, while O-QL and IQL achieve improvements of 80\% and 58\%, respectively. Similarly, the NDR-QL method yields superior path results, as also illustrated in Fig. \ref{fig:fig6}.

\begin{table*}[htbp]
    \centering
    \begin{tabularx}{\textwidth}{l : *{8}{Y}} 
    \hline\hline
    \multicolumn{1}{l}{}     & \multicolumn{8}{c}{Convergence steps $\downarrow$} \\ 
    \hline
    \multicolumn{1}{l}{MAPS} & 1       & 2       & 3       & 4       & 5       & 6       & 7        & 8       \\ 
    \hline
    QL                       & 252064  & 170762  & 82772   & 113510  & 51861   & 83744   & 170777   & 72285   \\
    O-QL \cite{zhou2024optimized}                    & 49656   & 33454   & 16560   & 25459   & 11918   & 15023   & 32382    & 13498   \\
    IQL \cite{low2022modified}                      & 98280   & 66788   & 64362   & 42316   & 45567   & 27130   & 23296    & 50907   \\
    NDR-QL                   & 27039   & 7676    & 7459    & 17524   & 8386    & 12154   & 14158    & 7649    \\ 
    \hline
    \multicolumn{1}{l}{~}    & \multicolumn{8}{c}{Shortest distance / Longest distance $\downarrow$} \\ 
    \hline
    QL                       & 213/395 & 103/484 & 103/649 & 116/390 & 112/571 & 98/386  & 136/1658 & 96/407  \\
    O-QL \cite{zhou2024optimized}                     & 174/322 & 84/390  & 100/462 & 125/262 & 120/165 & 98/364  & 98/208   & 98/192  \\
    IQL \cite{low2022modified}                     & 213/215 & 171/744 & 139/470 & 130/287 & 116/232 & 162/774 & 92/176   & 98/212  \\
    NDR-QL                   & 145/218 & 83/352  & 93/97   & 120/150 & 154/156 & 98/216  & 98/136   & 78/142  \\
    \hline\hline
    \end{tabularx}
    \caption{Comparison of the performance of different QL methods across eight map samples. The eight maps are illustrated in Fig. \ref{fig:fig2}, \ref{fig:fig4}, and \ref{fig:fig6}.}
    \label{tab:tab3}
\end{table*}

We further visualize the experimental results of MAP-1 and MAP-6. In Fig. \ref{fig:fig6}, the first row illustrates the number of steps per episode across algorithm iterations, while the second row depicts the total rewards per episode as the number of iterations increases.
We use gray dashed vertical lines to indicate the episode count at which the algorithm converges, further demonstrating the effectiveness of the NDR-QL method.
In the third row, we present the path results obtained by different methods, as well as the visualized Q-tables after convergence.
It can be observed that the output of the NDR model, serving as heuristic knowledge for QL, enables the algorithm to explore smaller regions, thereby accelerating the convergence process.

\subsection{Ablation Study}
In this subsection, we design two experiments to separately investigate the effectiveness of the NDR model and the NDR-QL method.

We examined the impact of different forms of model input data, as illustrated in Fig \ref{fig:fig1}, on model performance, with the results presented in Tab. \ref{tab:tab4}. In the table, `RGB' denotes the use of grayscale representations for maps with color-coded start and end points.
`Gaussian' refers to separating the channels of the start and end points from the map and encoding the start and end points using a Gaussian distribution. Additionally, we explore the impact of the Gaussian distribution parameters on performance.
Similarly to the Gaussian distribution, the sixth row in Tab. \ref{tab:tab4} investigates the effect of using the Euclidean distance to the start/end points as encoding values on performance.
Finally, `single pix' represents the use of a single pixel to denote the start/end points in their respective channels.

Additionally, we investigate the impact of the guideline-decoder on model performance, as shown in Fig \ref{fig:fig1}. In the table,   `GD' denotes the inclusion of the guideline-decoder module, while masking the GD module indicates that the model has a single output. This means that the model is trained using guideline labels to predict guidelines, and when predicting regions, the region is subsequently used as the label for supervised training.
We also use the F1-score to characterize the predictive performance of the model. Experiments conducted on standard datasets investigate the impact of each module on performance.

\begin{table}[!h]
    \centering
    \resizebox{\linewidth}{!}{%
    \begin{tabular}{lc;{1pt/1pt}cc;{1pt/1pt}cc} 
    \hline\hline
    \multicolumn{2}{c;{1pt/1pt}}{Settings ~} & \multicolumn{2}{c;{1pt/1pt}}{Seen} & \multicolumn{2}{c}{Unseen}  \\ 
    \hline
    Position Embeding        & GD            & R & G     & R     & G                   \\ 
    \hline
    RGB                      & \ding{55}             & 91.70            & 58.25 & 69.81 & 35.39               \\
    gaussian ($\sigma  = s/25$)  & \ding{55}             & 88.91            &55.34      & 67.87     & 33.71                   \\
    gaussian ($\sigma  = s/50$) & \ding{55}            & 91.19            & 57.02     & 69.98 & 37.65                   \\
    Euclidean dis            & \ding{55}             & 84.46                & 53.67     & 65.16    & 33.04                   \\
    single        pix        & \ding{55}             & 92.63            & 59.12 & 70.02 & 35.33               \\
    single        pix        & \ding{51}             & 93.02            & 61.22 & 71.96 & 37.31               \\
    \hline\hline
    \end{tabular}
    }
    \caption{Ablation study of the proposed NDR model}
    \label{tab:tab4}
    
\end{table}

As shown in Tab. \ref{tab:tab4}, the model's F1-score increases as the $\sigma$ parameter of the Gaussian distribution decreases. Ultimately, using single-pixel representations for start and end points outperforms the RGB-mixed input method, achieving the highest performance. The introduction of the Guideline Decoder further improves the model's performance by approximately 0.4\%.

In the second experiment, we investigate the effectiveness of using the guidelines and regions output by the NDR model to inform heuristic reward functions and heuristic Q-table initialization methods, respectively, and compare them with traditional distance-based heuristics.
We tested the convergence steps of different methods on five map samples, and the results are presented in Tab. \ref{tab:tab2}.
In the table, \textit{D-C} and \textit{D-Q} denote the distance-based heuristic continuous reward function and the Q-table initialization method, respectively.
\textit{N-C} and \textit{N-Q} represent the NDR model-based heuristic methods proposed in this paper, as described in Section \ref{subsec:reward} and Section \ref{subsec:q-table}.

Interestingly, the \textit{D-Q} method does not guarantee performance improvements for Q-learning. On maps like MAP-1, which contain many dead-ends, using the \textit{D-Q} method instead slows down the algorithm's convergence speed, resulting in 254794 versus 252064 convergence steps.
However, both \textit{N-C} and \textit{N-Q} methods can provide Q-learning with performance improvements ranging from 40\% to 90\%.
Combining the NDR heuristic methods with traditional distance-based heuristics achieves the highest convergence speed.

\begin{table}
    \centering

    \resizebox{\linewidth}{!}{%
    \begin{tabular}{cccc;{1pt/1pt}lllll} 
    \hline\hline
    \multicolumn{4}{c;{1pt/1pt}}{Settings} & \multicolumn{5}{c}{Scenarios}     \\ 
    \hline
    D-C & N-C & D-Q & N-Q       & MAP1 & MAP2 & MAP3 & MAP4 & MAP5  \\ 
    \hline
    \ding{51}      &         &         &               &150589      &28495      &66225      &83062      &44890       \\
                   &\ding{51}&         &               &23495      &26271      &12758      &32177      &33194       \\
    \ding{51}      &\ding{51}&         &               &21412      &22092      &12296      &31520      &25558       \\
                   &         &\ding{51}&               &254797      &77139      &60097      &51517      &86667       \\
                   &         &         &\ding{51}      &171416      &55262      &55224      &35153      &49353       \\
                   &         &\ding{51}&\ding{51}      &144470      &42049      &39654      &48542      &67215       \\
    \hdashline[1pt/1pt]
    \ding{51}      &\ding{51}&\ding{51}&\ding{51}      &27039      &7676     &7459      &17524      &8386       \\
    
    \hline\hline
    \end{tabular}
    
    }
    \caption{Ablation study of the proposed NDR-QL model on five map samples. The convergence steps of different settings are recorded.}
    \label{tab:tab2}
\end{table}

%% file: secs/conclusion.tex
\section{Conclusions}
\label{sec:limitation_conclusion}
In this work, we aimed to address the issue of slow convergence when applying traditional QL methods to path planning tasks.
Unlike previous methods that use distance as heuristic information, we propose that a pretrained neural network model can output a probability distribution of the optimal path, and using this distribution as heuristic information significantly accelerates the convergence of QL algorithms.
On the one hand, this paper follows the dual-output model previously proposed, which simultaneously predicts the region and guideline, and introduces improvements by separating the start and end points from the map’s channels while enhancing the fusion of high-level and low-level features. As a result, our redesigned NDR model outperforms the previous SOTA models on public datasets.
On the other hand, the proposed heuristic approach effectively leverages the region and guideline predictions output by the NDR model, performs reward prediction, and initializes the Q-table. Experimental results show that our proposed NDR-QL method achieves surprisingly fast convergence and surpasses previous improved QL methods.